\documentclass{article}

\usepackage[final]{neurips_2024}

\usepackage{natbib}
\setcitestyle{numbers,square}

\PassOptionsToPackage{numbers, compress}{natbib}

\usepackage[utf8]{inputenc} \usepackage[T1]{fontenc}    \usepackage{hyperref}       \usepackage{url}            \usepackage[table]{xcolor}

\usepackage{booktabs}       \usepackage{subcaption}
\usepackage{amsfonts}       \usepackage{nicefrac}       \usepackage{microtype}      \usepackage{xcolor}         
\usepackage{times}
\usepackage{epsfig}
\usepackage{amssymb}
\usepackage{mathtools}
\usepackage{enumitem}
\usepackage{setspace}
\usepackage{booktabs}
\usepackage{color,xcolor}
\usepackage{bbding}
\usepackage{diagbox}
\usepackage{bm}
\usepackage{multirow}
\usepackage{caption}
\usepackage{makecell}

\title{VLMimic: Vision Language Models are \\ Visual Imitation Learner for Fine-grained Actions}

\author{  Guangyan Chen$^1$ \And Meiling Wang$^1$ \And Te Cui$^1$ \And  Yao Mu$^2$ \And Haoyang Lu$^1$ \And Tianxing Zhou$^1$ \And  Zicai Peng$^1$ \And Mengxiao Hu$^1$ \And Haizhou Li$^1$ \And Li Yuan$^{3}$ \And Yi Yang$^{1}$ \thanks{Yufeng Yue and Yi Yang are co-corresponding authors. This work was  supported by the National Natural Science Foundation of China under Grant No. NSFC 62233002, 92370203. (email: yueyufeng@bit.edu.cn)} \And Yufeng Yue$^{1}$ \footnotemark[1] \AND
  $^1$ Beijing Institute of Technology \quad
  $^2$ The University of Hong Kong \quad
  $^3$ Peking University   
                                          }

\begin{document}

\maketitle

\begin{abstract}
 Visual imitation learning (VIL) provides an efficient and intuitive strategy for robotic systems to acquire novel skills. Recent advancements in Vision Language Models (VLMs) have demonstrated remarkable performance in vision and language reasoning capabilities for VIL tasks. Despite the progress, current VIL methods naively employ VLMs to learn high-level plans from human videos, relying on pre-defined motion primitives for executing physical interactions, which remains a major bottleneck. In this work, we present VLMimic, a novel paradigm that harnesses VLMs to directly learn even fine-grained action levels, only given a limited number of human videos. Specifically, VLMimic first grounds object-centric movements from human videos, and learns skills using hierarchical constraint representations, facilitating the derivation of skills with fine-grained action levels from limited human videos. These skills are refined and updated through an iterative comparison strategy, enabling efficient adaptation to unseen environments. Our extensive experiments exhibit that our VLMimic, using only 5 human videos, yields significant improvements of over 27\% and 21\% in RLBench and real-world manipulation tasks, and surpasses baselines by over 37\% in long-horizon tasks. Code and videos are available at \href{https://vlmimic.github.io/}{\textcolor{blue}{our home page}}.
\end{abstract}

\section{Introduction}
Visual Imitation Learning (VIL) has demonstrated remarkable efficacy in addressing various visual control tasks within intricate environments \cite{ mandlekar2020learning, tung2021learning, zeng2021transporter, brohan2022rt, brohan2023rt, padalkar2023open, chi2023diffusion,liang2023adaptdiffuser,huo2023human,liang2024skilldiffuser}. \textcolor{black}{Diverging from conventional approaches reliant on precise robot action labels, which often necessitates substantial human effort for data collection. Researchers increasingly turn to learning from human-object interaction videos that are easily accessible to reduce high data requirements. }

Existing methods for skill acquisition leveraging video data can be broadly categorized into two classes. One typical approach learns efficient visual representations for robotic manipulation through self-supervised learning from large volumes of videos\cite{grauman2022ego4d, xiao2022masked, nair2022r3m, shaw2023videodex, schmeckpeper2020reinforcement, edwards2019perceptual, schmeckpeper2020learning, chen2021learning, shao2021concept2robot, zakka2022xirl, das2021model，mu2023ec2}. Another approach focuses on learning task-relevant priors to guide robot behaviors or derive a heuristic reward function for reinforcement learning \cite{bahl2022human, shaw2023videodex, bahl2022human, sieb2020graph, sharma2019third, kumar2023graph, smith2019avid, liu2018imitation, xiong2021learning, mees2020adversarial, jang2022bc}. However, these approaches often encounter challenges when generalizing to unseen environments. Therefore, efficiently acquiring generalizable skills from limited videos remains highly challenging.

\textcolor{black}{An appealing prospect for handling this challenge is to employ large pretrained models by encapsulating extensive prior knowledge from broad data. Recent advances in vision-language models (VLMs) provide particularly promising tools in this regard, with their emergent and fast-growing conceptual understanding, commonsense knowledge, and reasoning abilities.} However, current VIL methods \cite{mu2024embodiedgpt,mu2024robocodex, chen2023human, wake2023gpt,patel2023pretrained, weng2024longvlm, li2023videochat, wang2024demo2code,sha2023languagempc,hu2023tree,gao2024dag} naively employ VLMs to learn high-level plans, and typically rely on a repertoire of pre-defined motion primitives. This reliance on individual skill acquisition is often considered a major bottleneck of the system due to the lack of large-scale robotic data. 
The question then arises: \textit{how can we leverage  VLMs to learn even fine-grained action levels directly from human videos, eliminating the reliance on predefined primitives?}

However, adapting  VLMs to achieve visual imitation learning for fine-grained actions is non-trivial due to the following critical reasons: 
(\uppercase\expandafter{\romannumeral1}) \textcolor{black}{Lack of fine-grained action recognition ability}. Despite existing advancements in VLMs, they still struggle to recognize low-level actions in videos. To overcome this obstacle, a human-object interaction grounding module is proposed, which parses videos into multiple segments, and estimates object-centric actions for subsequent analysis. Such that the intricate low-level action recognition task is converted into the pattern reasoning task,  which is more tractable for existing VLMs.
(\uppercase\expandafter{\romannumeral2}) Difficulty for VLMs in understanding motion signals.  
\textcolor{black}{Motion signals are characterized by inherent redundancy, hindering models from extracting valuable information.} 
To overcome this challenge, we propose hierarchical constraint representations for VLM reasoning, which exhibit semantic constraints through visualized actions and illustrate geometric constraints using keypoint values.
This representation effectively reduces redundancy and facilitates a comprehensive understanding, enabling our method to learn skills from a limited set of human videos.
(\uppercase\expandafter{\romannumeral3}) \textcolor{black}{Disparities in demonstration and target scenes}. Demonstration and execution scenes may involve different objects and tasks, impeding direct skill transfer. To this end, we propose a skill adapter with an iterative comparison strategy, which updates skills by iteratively contrasting with the demonstrated knowledge, facilitating the adaptation of learned skills to unseen scenes.

\begin{figure*}[t]
 \setlength{\abovecaptionskip}{-0.07cm}
    \begin{center}
    \includegraphics[width=14cm]{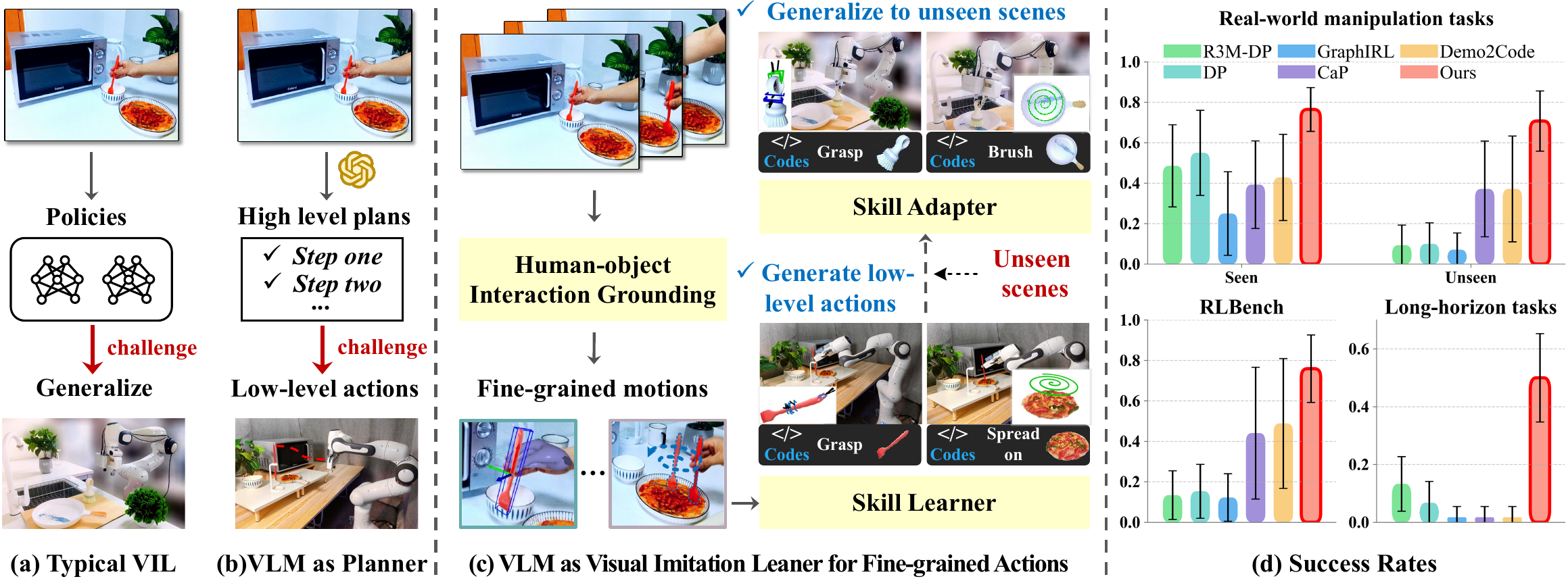}
    \end{center}
       \caption{{ Illustration of our VLMimic.}  \textcolor{black}{(a) Typical VIL methods struggle to generalize to unseen environments, and (b) current methods naively utilize VLMs as planners, encounter difficulties in generating low-level actions. (c) VLMimic grounds human videos to obtain action movements, and learns skills with fine-grained actions, while the skill adapter updates skills for generalization. (d) Our method achieves superior performance given a limited collection of human videos.} }
    \label{Figure1}
\vspace{-16pt}
\end{figure*}

Based on the above analysis, we present VLMimic, an approach that employs  VLMs to directly learn even fine-grained action levels from a limited number of human videos, and generalize to novel scenes. As shown in Fig. \ref{Figure1}, our method parses videos into multiple segments and captures object-centric movements using the human-object interaction grounding module. Then, a skill learner employing hierarchical constraint representations extracts knowledge from estimated motions, deriving skills with fine-grained actions. \textcolor{black}{In unseen environments}, a skill adapter with an iterative comparison strategy revises and updates the learned skills based on observations and task instructions.  
Extensive experiments demonstrate that VLMimic achieves strong performance across various scenes,  \textcolor{black}{utilizing only 5 human videos without requiring additional training}.

Our main contributions can be summarized as follows: (\uppercase\expandafter{\romannumeral1}) We propose VLMimic, a novel visual imitation learning framework empowered by VLMs, to learn generalizable robotic skills from human demonstration videos. VLMimic features a skill learner for knowledge extraction and a skill adapter for iterative skill refinement, \textcolor{black}{enabling efficient skill acquisition and adaptation}.
(\uppercase\expandafter{\romannumeral2}) We build an effective human-object interaction grounding algorithm to enhance fine-grained action recognition capabilities, and propose hierarchical constraint representations for VLM reasoning to reduce information redundancy and facilitate comprehensive action comprehension.
(\uppercase\expandafter{\romannumeral3}) Our method outperforms other methods by over 27\% on the RLBench. In real-world manipulation tasks, VLMimic achieves an improvement exceeding 21\% in seen environments and 34\% in unseen environments. Moreover, VLMimic exhibits an improvement of over 37\% in long-horizon tasks.

\section{Related Work}
\subsection{Learning from Human videos}
Conventional learning approaches necessitate access to expert demonstrations, which include observations and precise actions for each timestep. Drawing on human capabilities, learning from observation offers efficient and intuitive methods for robots to develop new skills. A plethora of recent researches explore leveraging large-scale human video data to improve robot policy learning \cite{grauman2022ego4d, xiao2022masked, nair2022r3m, shaw2023videodex, schmeckpeper2020reinforcement, edwards2019perceptual, schmeckpeper2020learning, chen2021learning, shao2021concept2robot, zakka2022xirl, das2021model}. Representative methods, R3M \cite{nair2022r3m} and MVP \cite{xiao2022masked}, which employ the internet-scale Ego4D dataset \cite{grauman2022ego4d} to pretrain visual representations for subsequent imitation learning tasks. Another thread of work \cite{bahl2022human, sieb2020graph, sharma2019third, kumar2023graph, smith2019avid, liu2018imitation, xiong2021learning, mees2020adversarial, jang2022bc} focuses on learning task-relevant priors from videos to guide robot behaviors or derive a heuristic reward function for reinforcement learning.
Learning by watching \cite{xiong2021learning} learns human-to-robot translation, the resulting representations are used to guide robots to learn robotic skills. 
WHIRL \cite{bahl2022human} infers trajectories and interaction details to establish a prior, but it learns policy through real-world exploration and requires a large number of rollouts to converge.
GraphIRL \cite{kumar2023graph} performs graph abstraction on the videos followed by temporal matching to measure the task progress, and a dense reward function is employed to train reinforcement learning algorithms. Despite these advancements, acquiring generalizable skills efficiently from limited demonstration videos remains highly challenging.

\subsection{Visual Imitation Learning with  VLMs}
Motivated by the notable success of  VLMs across various domains, recent research \cite{chen2023human, wake2023gpt,patel2023pretrained, weng2024longvlm, li2023videochat, wang2024demo2code} investigate their potential in VIL. GPT-4V for Robotics \cite{wake2023gpt} analyzes videos of humans performing tasks and outputs robot programs that incorporate insights into affordances. Digknow \cite{chen2023human} distills generalizable knowledge with a hierarchical structure, enabling the effective generalization to novel scenes. 
Demo2code \cite{wang2024demo2code} generates robot task code from demonstrations via an extended chain-of-thought and defines a common latent specification to connect the two. 
VLaMP \cite{patel2023pretrained} predicts visual planning from videos through video action segmentation and forecasting, handling long video history and complex action dependencies.
However, these approaches often rely on predefined movement primitives or pre-trained skills to execute lower-level actions, thereby only partially solving the control stack. In contrast, our investigation aims to push these boundaries and learn all lower-level actions for the robot, eliminating the reliance on predefined primitives and consequently broadening the applicability.

\section{VLMimic}
\begin{figure*}[t]
 \setlength{\abovecaptionskip}{-0.13cm}
    \begin{center}
    \includegraphics[width=0.95\textwidth]{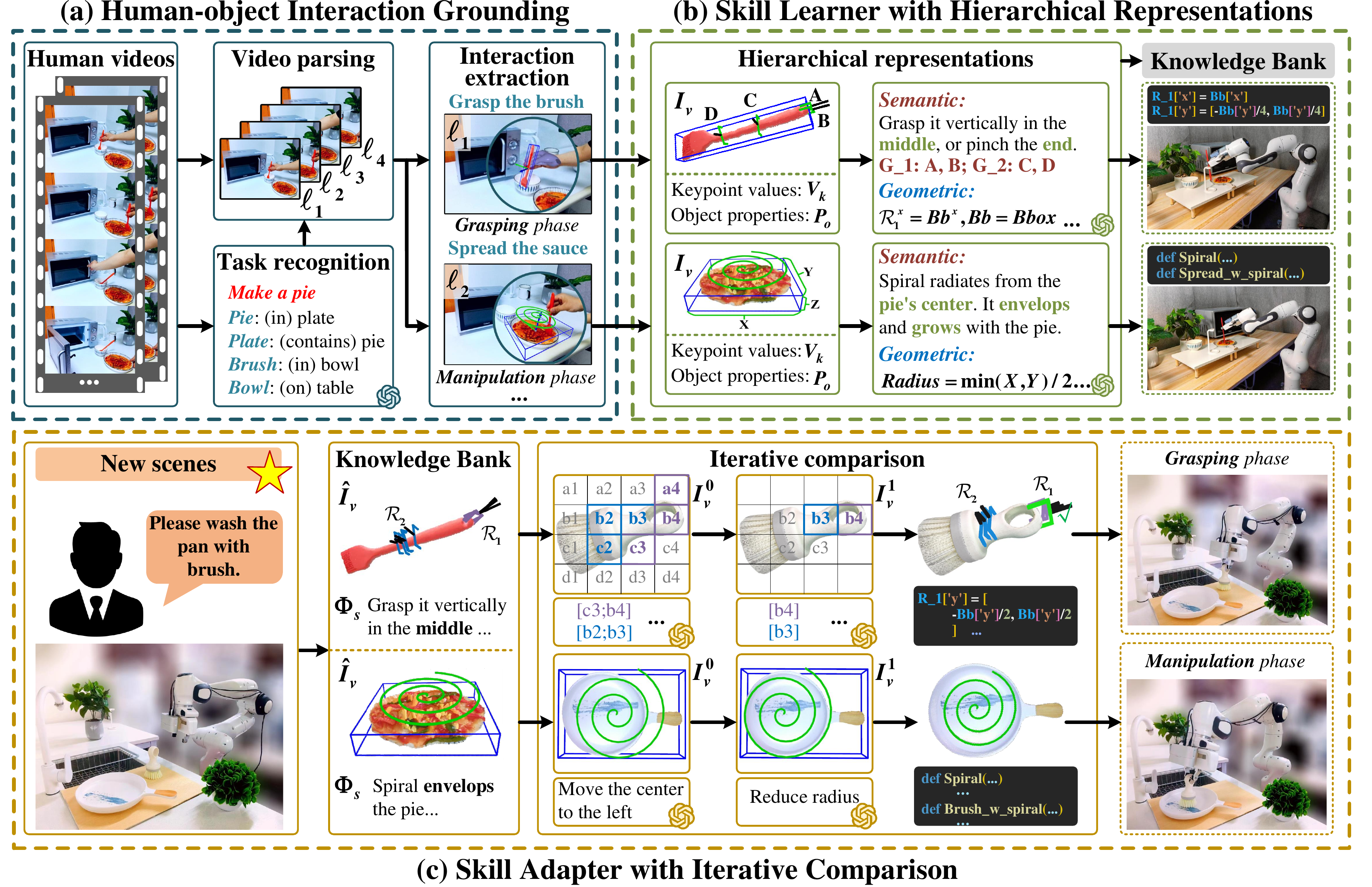}
    \end{center}
       \caption{{Illustration of our VLMimic. (a) The human-object interaction grounding module parses videos into multiple segments and captures object-centric movements. Then, (b) a skill learner extracts knowledge from action motions and derives skills. In novel scenes, (c) a skill adapter updates the learned skills to facilitate adaptation.}  }
    \label{Overview}
\vspace{-15pt}
\end{figure*}

Considering video demonstrations $\bm{\mathcal{V}}$ of a human performing manipulation tasks, recorded using an RGB-D camera. The overall pipeline of VLMimic is illustrated in Fig. \ref{Overview}. Our method first grounds human videos, segmenting them into subtask intervals $\{\bm{\tau}_i\}_{i=1}^{V} $ and capturing object-centric interactions $\bm{I}$. A skill learner with hierarchical representations then extracts knowledge from the obtained interactions, deriving skills with fine-grained actions. In unseen environments, a skill adapter employs an iterative comparison strategy to revise and update the learned skills based on observations and task instructions.

\subsection{Human-object Interaction Grounding}
\begin{figure*}[t]
 \setlength{\abovecaptionskip}{+0.07cm}
    \centering
\includegraphics[width=0.87\textwidth]{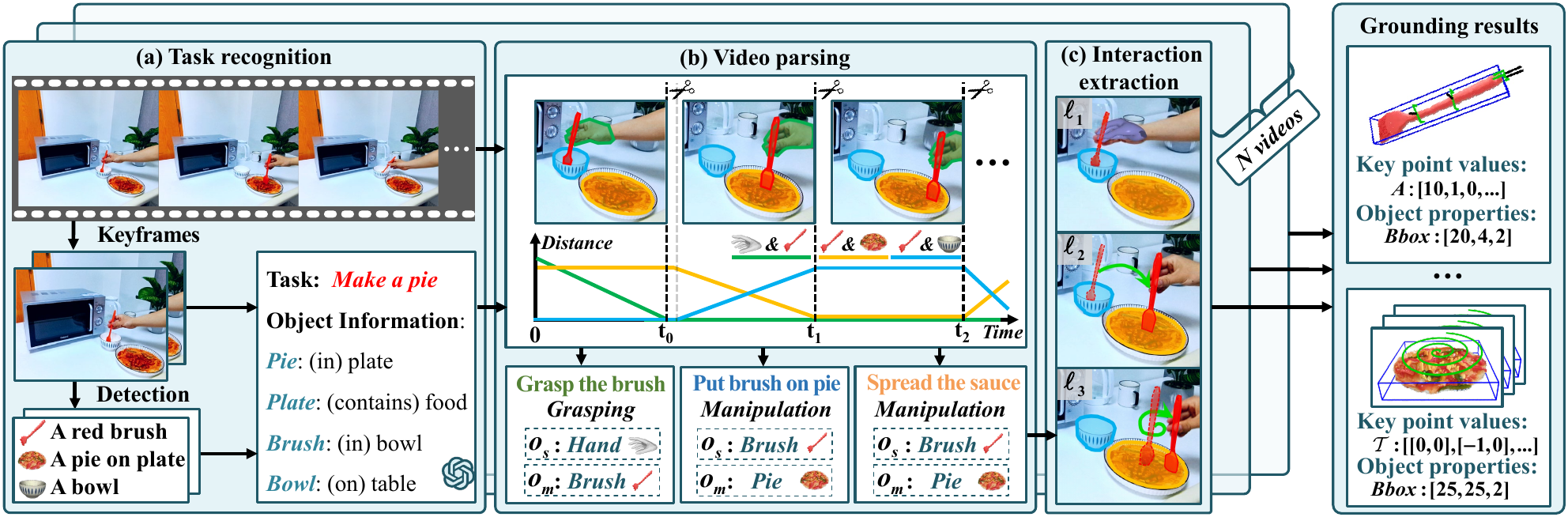}
           \caption{{ Illustration of Human-object interaction grounding module.} (a) It recognizes tasks and related objects from human videos, (b) parses videos into multiple segments based on this information, and subsequently (c) identifies object-centric interactions within each segment. }
       \vspace{-18pt}
    \label{Grounding}
\end{figure*}

Despite VLMs demonstrating proficiency in various vision tasks, they still struggle with fine-grained action recognition within videos. To mitigate this limitation, a four-stage process, illustrated in Fig. \ref{Grounding}, is utilized to extract object-centric interactions for skill learning, transforming this intricate problem into pattern reasoning problems, typically more tractable for existing VLMs.

\textbf{Task recognition}. Keyframes $\bm{\mathcal{K}}$ are intermittently extracted from videos $\bm{\mathcal{V}}$, vision foundation models ${\rm VFM}$ \cite{zhang2023recognize, wang2023caption, pan2023tokenize} are utilized to detect objects within these frames. Utilizing keyframes $\bm{\mathcal{K}}$ and textual detection results $\bm{T}_{d}$, VLMs are instructed to transcribe videos into task instructions $\bm{T}_t$, and compile the task-related objects $\bm{T}_o$ into textual information. The object information is predicated on the initial frame of the video data, comprising a list of object names and their spatial relationships. The task recognition procedure is formulated as follows:
\begin{equation} 
\setlength{\abovedisplayskip}{2pt}
\setlength{\belowdisplayskip}{2pt}
\begin{split}
       \bm{T}_{d} = {\rm VFM}(\bm{\mathcal{K}}), \quad \bm{T}_t, \bm{T}_o = {\rm VLM}(\bm{T}_{d}, \bm{\mathcal{K}}).
\end{split}
\end{equation}

\textbf{Video parsing}. Videos are parsed into segments $\{\bm{\tau}_i\}_{i=1}^{V} $, using interaction markers that identify interaction periods.
SAM-Track \cite{liu2023grounding, yang2021aot, yang2022deaot, kirillov2023segment, cheng2023segment} predicts hand and task-related object masks for each frame, 
and corresponding point clouds $\bm{\mathcal{P}}$ are generated through back-projection. Markers are then identified by determining the interaction start time $\bm{t}_i$ and end time $\bm{t}_e$, partitioning videos $\bm{\mathcal{V}}$ into multiple segments. \textcolor{black}{Segments with hand motion trajectory lengths below than $\gamma$ are filtered out}, yielding final set of segments $\{\bm{\tau}_i\}_{i=1}^{V} $. Concretely, the interaction markers are obtained as follows:
\begin{equation} 
\setlength{\abovedisplayskip}{2pt}
\setlength{\belowdisplayskip}{2pt}
\begin{split}
       \bm{d} = {\rm dist}(\bm{\mathcal{P}}), & \quad \bm{t}_i = \{t|\bm{d}^{t-1}\! >\! \epsilon \wedge \bm{d}^{t} < \epsilon\}, \quad \bm{t}_e = \{t|\bm{d}^{t-1}\! <\! \epsilon \wedge \bm{d}^{t} > \epsilon\},
       \end{split}
\end{equation}
where function ${\rm dist}$ calculates the distance between any two point clouds.

\textbf{Subtask recognition}. Each segment $\bm{\tau}_i$ is analyzed by VLMs, which generate a subtask textual description $\bm{T}_{\tau_i}$, and categorize the segment into grasping or manipulation phases based on the interacting entities and $\bm{T}_{\tau_i}$. VLMs also identify master objects $\bm{O}_{m}$ and slave objects $\bm{O}_{s}$. In the grasping phase, the agent performs a reach-and-grasp action targeting $\bm{O}_{m}$, designating the hand as $\bm{O}_{s}$. In the manipulation phase, the agent employs $\bm{O}_{s}$ to interact with $\bm{O}_{m}$. 

\textbf{Object-centric interaction extraction}. FrankMocap \cite{rong2020frankmocap} and the Iterative Closest Point (ICP) algorithm \cite{besl1992method, rusinkiewicz2001efficient} are employed to derive precise hand pose trajectories, which are subsequently converted into robot gripper pose trajectories. Furthermore, BundleSDF \cite{wen2023bundlesdf} is employed for object reconstruction, and FoundationPose \cite{wen2023foundationpose} is leveraged for object pose estimation based on reconstructed objects ${\bm O}$. In grasping phases, interactions $\bm{I}$ are represented as grasp poses at hand-object contacts. For manipulation phases, $\bm{I}$ are defined as trajectories of slave objects $\bm{O}_s$ relative to master objects $\bm{O}_m$. 
This object-centric paradigm facilitates efficient skill acquisition and enables VLMimic to accommodate demonstrations across diverse viewpoints.

\subsection{Skill Learner with Hierarchical Representations} \label{learner}

A straightforward approach for learning skills involves directly discerning the numerical trajectory patterns \cite{wang2023prompt, mirchandani2023large}. However, VLMs face challenges in reasoning about inherently redundant motion signals, limiting their ability to extract valuable information. To reduce redundancy and foster comprehensive comprehension, hierarchical constraint representations are proposed for skill learning, as illustrated in Fig. \ref{Overview}. These representations exhibit semantic constraints via visualized interaction $\bm{I}_v$ and further detail the fine-grained geometric constraints by integrating keypoint values $\bm{V}_k$.

\textbf{Learning with hierarchical constraint representations}. \textcolor{black}{Rendering} interaction $\bm{I}$, and textual notations $\bm{T}_n$ on objects ${\bm O}$ to derive visualized interaction $\bm{I}_{v}$, VLMimic facilitates reasoning capabilities to analyze semantic constraints $\bm{\Phi}_{s}$ by encouraging VLMs to attend to objects and their related actions, and integrating keypoint values $\bm{V}_k$ and object properties $\bm{P}_o$ (e.g., 3-D bounding boxes) to derive geometric constraints $\bm{\Phi}_{g}$. Formally, constraints are learned as follows: 
 \begin{equation} 
\setlength{\abovedisplayskip}{1pt}
\setlength{\belowdisplayskip}{1pt}
\begin{split}
       \bm{I}_v = {\rm Render}([\bm{I}, \bm{T}_n],\bm{O}), \quad \bm{\Phi}_{s} = {\rm {S}_l}(\bm{I}_v) \quad  \bm{\Phi}_{g} = {\rm {G}_l}(\bm{\Phi}_{s}, \bm{V}_k, \bm{P}_o),
\end{split}
\end{equation}
where ${\rm {S}_l}$, and ${\rm {G}_l}$ are functions to learn semantic, and geometric constraints, respectively.

(\uppercase\expandafter{\romannumeral1}) Grasping constraints. Inspired by task space regions (TSRs) \cite{berenson2011task}, the grasping constraint $\bm{\Phi}_{g}$ can be approximated as a series of bounded regions $\{\bm{\mathcal{R}}_i\}_{i=1}^{N_C}$.
\textcolor{black}{Interactive grasp poses} $\bm{I}$ are exhibited on objects, each associated with an index notation $\bm{T}_n$. These visualized interactions $\bm{I}_v$ are presented to VLMs, leveraging their inherent knowledge and visual understanding ability to summarize semantic constraints $\bm{\Phi}_{s}$ and group these poses. Geometric constraints $\bm{\Phi}_{g}$, represented as bounded regions, are derived by calculating ranges of grasp pose values $\bm{V}_k$ within the same group, and associating them with object properties $\bm{P}_{o}$.
This approach simplifies the complex task of constraint region generation into a series of visual understanding based multiple-choice question answering. 
Moreover, representing constraints through object properties enhances generalization across objects.

(\uppercase\expandafter{\romannumeral2}) Manipulation constraints. Interaction trajectory $\bm{I}$ is delineated on the master object ${\bm{O}_m}$, incorporating keypoints $\bm{V}_k$ in the textual prompt. Semantic constraints $\bm{\Phi}_{s}$ are identified by VLMs based on the visualized interaction $\bm{I}_v$ and subtask description $\bm{T}_{\tau_i}$. Geometric constraints $\bm{\Phi}_{g}$ are then formulated based on semantic constraints $\bm{\Phi}_{s}$, keypoint values $\bm{V}_k$, and object properties $\bm{P}_o$, expressing $\bm{\Phi}_{g}$ via the trajectory code. \textcolor{black}{The code comprises two components: parameter estimation functions $\bm{f}_p$, which derives trajectory parameters from object properties, and trajectory generation functions $\bm{f}_s$, employing estimated parameters to generate a sequence of slave object poses relative to the master object, promoting effective generalization across various objects and spatial configurations}. 

\textcolor{black}{During execution, grasp candidates are uniformly sampled within the learned grasping constraints, and object-centric trajectories predicted from manipulation constraints are converted to end-effector trajectories in the world frame using each grasp candidate of the slave object and object poses of master and slave objects. The resulting end-effector trajectory candidates are evaluated using motion planner, such as OMPL \cite{sucan2012open}, the trajectory with the highest fraction is selected.}

\textbf{Knowledge bank construction}. A knowledge bank $\bm{B}$ is established to archive both high-level planning and low-level skill insights, storing knowledge with key-value pairs $(\bm{k}_i, \bm{v}_i)$. 
High-level planning knowledge is indexed using task description $\bm{T}_t$ as keys, paired with the consequent action sequence $\bm{T}_{\tau}$ as values. For low-level skill knowledge, keys are constituted by the object images and subtask description $\bm{T}_{\tau_i}$, and values comprise reconstructed objects, as well as semantic constraints $\bm{\Phi}_s$ and geometric constraints $\bm{\Phi}_g$ representing learned skills.

\subsection{Skill Adapter with Iterative Comparison} 
Even though the skill learner exhibits efficient skill acquisition, the demonstration and execution scenes may differ in objects and tasks, impeding direct skill transfer to unseen environments. To mitigate these challenges,  VLMs are instructed to adapt skills via an iterative comparison strategy, as depicted in Fig. \ref{Overview}. This approach updates learned skills by iteratively contrasting with the demonstrated knowledge, thereby enabling effective adaptation of retrieved skills to novel scenes. 

\textbf{High level planning}. High-level planning knowledge $\bm{T}_{\tau}$ is retrieved from knowledge bank $\bm{B}$ based on the task instruction, which acts as the in-context example for  VLMs, along with the scene observation.  VLMs serve as a physically-grounded task planner \cite{skreta2024replan, hu2023look}, generating a sequence of actionable steps and descriptions of task-related objects $\bm{T}_{o}$.

\textbf{Iterative comparison}. 
In each iteration,  VLMs perform a comparative analysis between the adapted interaction $\bm{I}$ and retrieved interaction $\bm{\hat{I}}$, subsequently updating the skill constraints $\bm{{\Phi}}_{s}$ and $\bm{{\Phi}}_{g}$. This iterative process persists until either convergence is achieved or the maximum number of iterations $N_I$ is reached. This approach facilitates reasoning in  VLMs by directing their attention to discrepancies, and enables VLMs to pinpoint the best available solution through an iterative process. The adapting procedure at the $i$-th iteration can be formally represented as:
\begin{equation} 
\setlength{\abovedisplayskip}{2pt}
\setlength{\belowdisplayskip}{2pt}
\begin{split}
\bm{{I}}_v^{i} = {\rm Render}(\bm{{\Phi}}_{g}^{i}, O), \quad
\bm{{\Phi}}_{s}^{i\!+\!1} = {\rm {S}_a}(\bm{\hat{I}}_v, \bm{{I}}_v^{i},\bm{{\hat{\Phi}}}_{s}, \bm{{\Phi}}_{s}^{i}), \quad  \bm{{\Phi}}_{g}^{i\!+\!1} = {\rm {G}_a}( \bm{{\hat{\Phi}}}_{g}, \bm{{\Phi}}_{g}^{i}, \bm{{\Phi}}_{s}^{i\!+\!1}, \bm{V}_k, \bm{P}_o),
\end{split}
\end{equation}
where $\bm{{\hat{\Phi}}}_{g}$ and $\bm{{\hat{\Phi}}}_{s}$ denote referential constraints, extracted from the knowledge base. The functions ${\rm {S}_a}$ and ${\rm {G}_a}$ adapt semantic and geometric constraints, respectively.

(\uppercase\expandafter{\romannumeral1}) Grasping constraint adaptation. 
As the grasping orientation is typically derivable from position constraints using grasping models \cite{fang2020graspnet, fang2023anygrasp, ju2024robo}, our work focuses on transferring position constraints.
The visualized grasping position space is discretized into an $m \times n$ grid ($m, n \in Z$) and annotated with textual notations $\bm{T}_n$, obtaining $\bm{{I}}_v^{0}$. VLMs are instructed to update semantic constraints $\bm{{\Phi}}_{s}$, by contrasting with the referential interaction $\bm{\hat{I}}_v$ and semantic constraints $\bm{{\hat{\Phi}}}_{s}$, 
and to adapt geometric constraints $\bm{{\Phi}}_{g}$ by sampling $K$ outputs of grasping region selection. The updated $\bm{{\Phi}}_{g}$ are then visualized for the next iteration. 
The 3-D positional region is represented using two perspectives, and the consistency of the selected regions for the overlapping area validates the VLM outputs. 
The obtained constraints are expressed via object properties to enhance generalization.

(\uppercase\expandafter{\romannumeral2}) Manipulation constraint adaptation. VLMs are instructed to iteratively self-summarise and update manipulation constraints based on the task instruction and scene differences. 
VLMimic generates trajectories adhering to geometric constraints $\bm{\Phi}_g$, which are exhibited on master objects. VLMs are instructed to analyze the deviation of the adapted interaction $\bm{{I}}_v$ from the referential interaction $\bm{\hat{I}}_v$ to revise semantic constraints $\bm{\Phi}_s$, and geometric constraints $\bm{\Phi}_g$ undergo refinement predicated on the updated $\bm{\Phi}_s$, along with trajectory keypoint values $\bm{V}_k$ and object properties $\bm{P}_o$.

\textbf{Failure reasoning}. Despite the ability of  VLMs to generate effective constraints, environmental noise, such as trajectory estimation errors, impedes successful task execution. Thus, we leverage  VLMs to detect and address failures during execution by providing them with perceptual results, such as object pose and robot end-effector trajectories, enabling autonomous failure identification and reasoning for rectification.

\begin{table}[t]
\caption{Success rates on RLbench. "Obs-act", "Template", and "Video" indicate paired observation-action sequences, code templates, and videos performing subtasks.}
\label{RLBench}
\begin{minipage}{\textwidth}
\makeatletter\def\@captype{table}
\begin{subtable}[t]{\textwidth}
\resizebox{\linewidth}{!}{
\fontsize{8}{10}\selectfont
\begin{tabular}{>{\centering\arraybackslash}m{2.06cm} | *{7}{>{\centering\arraybackslash}m{1.49cm}}}
\toprule[1.5pt]

Methods& {R3M-DP} & {DP}   & {GraphIRL} & {CaP}                      & {Demo2Code} & {Ours}                     \\
\midrule
Overall & $0.13 (\pm 0.12)$ & $0.15 (\pm 0.13)$ & $0.12 (\pm 0.12)$ & $0.44 (\pm 0.33)$ & $0.49 (\pm 0.32)$ & $\bm{0.76 (\pm 0.17)}$ \\
\bottomrule
\end{tabular}
}
\end{subtable}
\makeatletter\def\@captype{table}
\begin{subtable}[t]{\textwidth}
\centering
\resizebox{\linewidth}{!}
{\fontsize{8}{10}\selectfont
\begin{tabular} {@{}lcccccccccccc@{}}
\toprule
Methods & \thead{Type of\\demos } & \thead{Num of\\demos}&   \thead{Reach\\ target}  &   \thead{Take lid off\\ saucepan} & 
                  \thead{Pick\\ up cup} & 
                  \thead{Toilet\\seat up} & 
                  \thead{Open\\box} & 
                  \thead{Open\\door} \\
\midrule
   R3M-DP & Obs-act & 100 & 0.37 & 0.20 & 0.20 & 0.07 & 0.02 & 0.25\\ 
        DP & Obs-act & 100 & 0.43 & 0.25 & 0.24 & 0.05 & 0.04 & 0.22 \\ 
        GraphIRL & Video & 100 &0.39 & 0.14 & 0.23 & 0.03 & 0.03 & 0.21\\ 
        CaP & Template & 5 &   {0.95} &  {0.90} & 0.58 & 0.05 & 0.12 & 0.65\\ 
        Demo2Code & Video & 5 &0.94 & 0.86 &  {0.65} & 0.06 &  {0.19} &  {0.83} \\ 
        \rowcolor{gray!25} \textbf{Ours} & \textbf{Video} & \textbf{5} & \textbf{0.97} & \textbf{0.94} & \textbf{0.80} & \textbf{0.76} & \textbf{0.75} & \textbf{0.90}\\ 
\toprule
{Methods} & \thead{Type of\\demos } & \thead{Num of\\demos}& 
                  \thead{Meat off \\ grill} &\thead{Open\\drawer} & 
                  \thead{Open\\grill} & 
                  \thead{Open\\microwave} & 
                  \thead{Open\\oven} & 
                  \thead{Knife on\\board}\\
\midrule
    R3M-DP & Obs-act & 100 & 0.15 & 0.25 & 0.07 & 0.03 & 0.00 & 0.00\\ 
        DP & Obs-act & 100 & 0.17 & 0.28 & 0.09 & 0.07 & 0.00 & 0.00 \\ 
        GraphIRL & Video & 100 &0.16 & 0.18 & 0.04 & 0.04 & 0.02 & 0.00 \\ 
        CaP & Template & 5 &0.35 & 0.17 & 0.46 & 0.12 & 0.16 & 0.78 \\ 
        Demo2Code & Video & 5& {0.57} &  {0.22} &  {0.40} &  {0.14} &  {0.21} &  {0.79}\\  
        \rowcolor{gray!25} \textbf{Ours} & \textbf{Video} & \textbf{5} & \textbf{0.79} & \textbf{0.75} & \textbf{0.81} & \textbf{0.45} & \textbf{0.43} & \textbf{0.76} \\ 
\bottomrule[1.5pt]
\end{tabular} 
}
\end{subtable}
\end{minipage}
\vspace{-16pt}
\end{table}

\section{Experiments} \label{Exp.}

\textbf{Baselines}. VLMimic is compared with five representative methods: (1) R3M-DP that utilizes the pre-trained R3M visual representation \cite {nair2022r3m} with the state-of-the-art (SOTA) diffusion policy \cite{chi2023diffusion}; (2) Diffusion Policy (DP) \cite{chi2023diffusion}, a SOTA end-to-end policy method;  (3) GraphIRL \cite{kumar2023graph}, a method that employs graph abstraction and learns reward functions for reinforcement learning (RL); (4) Code as Policy (CaP) \cite{liang2023code}, an LLM-driven method that re-composes API calls to generate new policy code; and (5) Demo2code \cite{wang2024demo2code}, an LLM-driven planner method that translates demonstrations into task code. We modify it to integrate the analysis results from GPT-4V for Robotics \cite{wake2023gpt}, enabling it to transcribe videos into code.
R3M-DP and DP are trained using the robot demonstrations with paired observation and action sequences. GraphIRL is trained in simulators with paired robot videos, Demo2code and our method learns skills with human videos in real-world experiments and robot videos in simulation experiments.

\subsection{Simulation Manipulation Tasks}
\textbf{Experimental setup}. To assess our approach on challenging robotic manipulation tasks, the RLBench \cite{james2020rlbench} benchmark is utilized for simulation tasks. Due to the unavailability of human videos in simulations, demo2code and our method utilize robot videos captured from a single-camera perspective during demonstrations, incorporating robot gripper trajectories.

\textbf{Results}. We investigate the capacity of VLMimic to acquire skills from a limited collection of video demonstrations, without requiring additional training. Our evaluation encompasses 12 manipulation tasks, as detailed in Table \ref{RLBench}, demonstrating that our method surpasses all other methods in 11 out of these tasks. Our method, learned with only 5 human videos, obviously outperforms R3M-DP and DP by over 61\% in overall performance, despite both being trained on 100 robot demonstrations. Compared to CaP and demo2code, our method demonstrates an improvement exceeding 27\%, highlighting the significant performance enhancements facilitated by the VLMimic framework.

\subsection{Real-world Manipulation Tasks}
\textbf{Experimental setup}. The real-world testing environment (E) is divided into "seen" (SE) and "unseen" (UE) categories. The "seen" category allows for testing in the environment where demonstrations were collected, whereas the "unseen" category involves testing in a distinct environment characterized by different objects and layouts. Success criteria are human-evaluated and the success rate is calculated from 10 randomized object positions and orientations.

\textbf{Results}: To validate the effectiveness of VLMimic in real-world settings, we conduct experiments involving 14 challenging real-world manipulation tasks selected from recent robotics research \cite{ahn2022can, xiao2022robotic, brohan2022rt, yu2023scaling}.
Quantitative results, presented in Table \ref{Real_mani}, demonstrate that VLMimic clearly outperforms other methods across all tasks, particularly in the "unseen" environment (UE). VLMimic achieves an improvement exceeding 21\% in SE and more than 34\% in UE. Results reveal the outstanding ability of VLMimic to acquire skills from human videos and adapt them to unseen environments.

\subsection{Real-world Long-Horizon Tasks}
\textbf{Experimental setup}. \textcolor{black}{Since baseline methods struggle to complete long-horizon tasks in the UE setting, experiments are conducted in the SE setting. All other experimental settings are consistent with those in the real-world manipulation task.}

\textbf{Results}. The performance of VLMimic on long-horizon tasks is quantitatively evaluated by its successful completion of six distinct tasks, each comprising at least five subtasks.
Experimental results, as depicted in Table \ref{Long}, obviously exhibit a substantial enhancement achieved by our method over baseline methods. These outcomes suggest that the proposed method is capable of developing robust skills, thereby achieving promising performance in even long-horizon tasks.

\begin{table}[t]
\caption{Success rates on real-world manipulation experiments. "Obs-act", "Template", and "Video" indicate paired observation-action sequences, code templates, and videos performing subtasks. "SE" and "UE" are seen and unseen environments. }
\label{Real_mani}
\begin{minipage}{\textwidth}

\makeatletter\def\@captype{table}
\begin{subtable}[t]{\textwidth}
\resizebox{\linewidth}{!}{
{\fontsize{8}{10}\selectfont
\begin{tabular}{>{\centering\arraybackslash}m{2.5cm} | *{7}{>{\centering\arraybackslash}m{1.49cm}}}
\toprule[1.5pt]

Methods& R3M-DP & DP   & GraphIRL & CaP                      & Demo2Code & Ours                    \\
\midrule
Overall (SE) & $0.49 (\pm 0.20)$ & $0.55 (\pm 0.21)$ & $0.25 (\pm 0.21)$ & $0.39 (\pm 0.22)$ & $0.43 (\pm 0.21)$ & $\bm{0.76 (\pm 0.11)}$                     \\
Overall (UE) & $0.09 (\pm 0.10)$ & $0.10 (\pm 0.10)$ & $0.07 (\pm 0.08)$ & $0.37 (\pm 0.24)$ & $0.37 (\pm 0.26)$ & $\bm{0.71 (\pm 0.15)}$ \\

\bottomrule
\end{tabular}
}
}
\end{subtable}

\makeatletter\def\@captype{table}
\begin{subtable}[t]{\textwidth}
\centering
\resizebox{\linewidth}{!}{
{\fontsize{8}{10}\selectfont
\begin{tabular}{lcccccccccccccccccccc}
\toprule
\multirow{2}{*}{Methods} & \multirow{2}{*}{\thead{Type of\\demos }} & \multirow{2}{*}{\thead{Num of\\demos}} & \multicolumn{2}{c}{\thead{Open\\drawer}} & \multicolumn{2}{c}{\thead{Stack\\block}} & \multicolumn{2}{c}{\thead{Open\\oven}} &\multicolumn{2}{c}{\thead{Put fruit \\ on plate}} & \multicolumn{2}{c}{\thead{Press\\button}} & \multicolumn{2}{c}{\thead{Open\\microwave}} & \multicolumn{2}{c}{\thead{Put tray\\in oven}}\\
\cmidrule(lr){4-5} \cmidrule(lr){6-7} \cmidrule(lr){8-9} \cmidrule(lr){10-11} \cmidrule(lr){12-13} \cmidrule(lr){14-15}
\cmidrule(lr){16-17} 
& & & SE & UE & SE & UE  & SE & UE &SE & UE  & SE & UE  & SE & UE  & SE & UE \\

\midrule
R3M-DP & Obs-act & 100 & 0.2 & 0.1 &  {0.6} & 0.2 & 0.3 & 0.0 &  {0.8} & 0.3 &  {0.7} & 0.2 & 0.2 & 0.0 &  {0.4} & 0.0 \\
DP & Obs-act & 100 & 0.3 & 0.1 &  {0.6} & 0.2 &  {0.4} & 0.1 &  {0.9} & 0.4 &  {0.7} & 0.1 & 0.3 & 0.0 &  {0.4} & 0.0 \\
GraphIRL & Video & 100 & 0.2 & 0.0 & 0.4 & 0.1 & 0.0 & 0.0 & {0.7} & 0.2 & 0.4 & 0.2 &  {0.0} &  {0.0} & 0.2 & 0.0 \\
CaP & Template & 5 & 0.3 &  {0.3} & 0.5 &  {0.5} & 0.3 &  {0.2} & 0.8 &  {0.8} &  {0.7} & 0.7 & 0.1 &  {0.1} & 0.2 & 0.1 \\
Demo2Code & Video & 5 &  {0.3} & 0.3 & 0.5 & 0.4 & 0.3 & 0.1 & 0.8 &  {0.9} & {0.8} &  {0.8} & 0.2 &  {0.1} & 0.3 &  {0.2} \\
\rowcolor{gray!25} Ours & Video & 5 & \textbf{0.8} & \textbf{0.7} & \textbf{0.9} & \textbf{0.8} & \textbf{0.6} & \textbf{0.6} & \textbf{0.9} & \textbf{0.9} & \textbf{0.8} & \textbf{0.9} & \textbf{0.7} & \textbf{0.6} & \textbf{0.7} & \textbf{0.7} \\
\toprule
    \multirow{2}{*}{Methods} & \multirow{2}{*}{\thead{Type of\\demos }} & \multirow{2}{*}{\thead{Num of\\demos}}  & \multicolumn{2}{c}{\thead{Turn on\\oven}} & \multicolumn{2}{c}{\thead{Sweep\\table}} & \multicolumn{2}{c}{\thead{Insert\\box}} & \multicolumn{2}{c}{\thead{Brush\\pan}} & \multicolumn{2}{c}{\thead{Sauce\\spread}} & \multicolumn{2}{c}{\thead{Put toy\\to drawer}} & \multicolumn{2}{c}{\thead{Pour from\\cup to cup}}\\
    \cmidrule(lr){4-5} \cmidrule(lr){6-7} \cmidrule(lr){8-9} \cmidrule(lr){10-11} \cmidrule(lr){12-13} \cmidrule(lr){14-15}
\cmidrule(lr){16-17}
&&&SE&UE&SE&UE&SE&UE&SE&UE&SE&UE&SE&UE&SE&UE \\
\midrule
                        R3M-DP & Obs-act & 100 & 0.2 & 0.0 & 0.7 & 0.2 &  {0.4} & 0.0 & 0.6 & 0.1 & 0.6 & 0.1 & 0.6 & 0.1 &  {0.5} & 0.0 \\
DP & Obs-act & 100 &  {0.3} & 0.0 &  {0.8} & 0.1 & 0.3 & 0.1 &  {0.7} & 0.1 &  {0.7} & 0.0 &  {0.7} & 0.1 & {0.6} & 0.1 \\
GraphIRL & Video & 100 &  {0.2} & 0.1 & 0.5 & 0.2 & 0.0 & 0.0 & 0.2 & 0.0 & 0.2 & 0.1 & 0.4 & 0.1 & 0.1 & 0.0 \\
CaP & Template & 5 &  {0.3} &  {0.3} & 0.6 & 0.5 & 0.1 & 0.1 & 0.3 &  {0.4} & 0.3 & 0.3 & 0.6 &  {0.7} & 0.4 &  {0.2} \\
Demo2Code & Video & 5 & 0.2 & 0.1 & 0.6 &  {0.6} & 0.3 &  {0.2} & 0.4 & 0.3 & 0.3 &  {0.4} &  {0.7} & 0.6 & 0.3 &  {0.2} \\
\rowcolor{gray!25}Ours & Video & 5 & \textbf{0.8} & \textbf{0.7} & \textbf{0.9} & \textbf{0.9} & \textbf{0.6} & \textbf{0.4} & \textbf{0.8} & \textbf{0.7} & \textbf{0.8} & \textbf{0.7} & \textbf{0.8} & \textbf{0.8} & \textbf{0.6} & \textbf{0.5} \\
\bottomrule[1.5pt]
\end{tabular}
}
}
\end{subtable}
\end{minipage}
\vspace{-13pt}
\end{table}

\begin{table}[t]
\caption{Success rates on long-horizon tasks. "Obs-act", "Template", and "Video" indicate observation-action sequences, code templates, and videos performing tasks. }
\label{Long}
\resizebox{\linewidth}{!}{
{\fontsize{8}{10}\selectfont
\begin{tabular}{lccccccccc}
\toprule[1.5pt]
Methods & \thead{ Type of \\ demos} & \thead{ Num of \\ demos} &\thead{  Make \\ coffee} & \thead{ Clean \\ table} & \thead{ Make \\ a pie} & \thead{ Wash \\ pan} & \thead{ Make \\slices} & \thead{ Chem.\\ exp.} & \thead{Overall } \\
\midrule
   R3M-DP & Obs-act & 100 & 0.10 &  {0.30} &  {0.20} &  {0.10} &  {0.00} &  {0.10} &  $0.13 (\pm 0.09)$ \\
    DP & Obs-act & 100 & 0.00 & 0.20 & 0.10 & 0.00 & 0.10 & 0.00 & $0.07 (\pm 0.07)$ \\
    GraphIRL& Video & 100 & 0.00 & 0.10 & 0.00 & 0.00 & 0.00 & 0.00 & $0.02 (\pm 0.04)$  \\
    CaP& Template & 5 & 0.00 & 0.10 & 0.00 & 0.00 & 0.00  & 0.00 & $0.02 (\pm 0.04)$ \\
    Demo2Code& Video & 5 &0.00 &0.10 & 0.00& 0.00& 0.00&  0.00  & $0.02 (\pm 0.04)$\\
    \rowcolor{gray!25} \textbf{Ours}& \textbf{Video} & \textbf{5} & \textbf{0.40} & \textbf{0.70} & \textbf{0.70} & \textbf{0.40} & \textbf{0.50} & \textbf{0.30} & $\bm{0.50 (\pm 0.15)}$  \\
\bottomrule[1.5pt]
\end{tabular}
}
}
\vspace{-16pt}
\end{table}

\subsection{Ablation Studies}
Comprehensive ablation studies are conducted to investigate the fundamental designs of our VLMimic approach. The effects of these design decisions are assessed by measuring the success rate on real-world manipulation tasks, which is computed across 10 randomized object positions and orientations.

\textbf{Hierarchical constraint representations}. 
Table \ref{Ablation} (a) analyzes three distinct constraint representations. Variants that exclusively reason semantic constraints or directly obtain geometric constraints without semantic analysis, lead to diminished performance. The results exhibit that hierarchical constraint representations enhance skill acquisition capabilities, demonstrating the pivotal role in facilitating the understanding and reasoning capabilities of VLMs.

\textbf{Grasping learning}. Table \ref{Ablation} (b) presents variants and their respective performance. The first variant utilizes VLMs for direct prediction of constraint region values, resulting in a significant performance decline. The second variant employs the DBScan clustering algorithm to group grasp poses and derive constraints as bounded regions. However, this method only considers numerical distributions without incorporating grasping common sense, leading to performance degradation.

\textbf{Number of human videos}. Table \ref{Ablation}(c) presents an analysis of the impact of human video quantity on performance. Results indicate that our method attains high success rates on complex tasks with a single human video demonstration, and increasing the number of videos yields performance gains. The results show that our approach can efficiently learn generalizable skills from a limited number of human videos. We choose to use 5 demonstration videos to balance data availability and performance.

\textbf{Comparison strategy}. Table \ref{Ablation} (d) analyzes the impact of the comparison strategy in skill adapters. Variants compare constraints exclusively utilizing either visualized interactions or keypoints exhibit decreased success rates. Visual comparison facilitates semantic contrast in VLM, while keypoint values provide fine-grained geometric information. Experimental results illustrate that our strategy facilitates reasoning for both semantic and geometric constraint adaptation.

\textbf{Number of iterations}. We conduct an analysis on the impact of iteration count in skill adapter and search for the optimal choice, as shown in Table \ref{Ablation}(e). Reducing the number of iterations to 0 results in a noticeable decrease in performance. Strong results are observed in the initial iteration, with modest improvements in subsequent iterations.  The findings indicate that this iterative approach enhances the effectiveness of skill adaptation by enabling VLMs to identify the best available solution. For enhanced performance, 4 iterations are selected.

\textbf{Failure reasoning}. The impact of failure reasoning is investigated in Table \ref{Ablation}(f). The success rate exhibits an upward trend with increasing iterations, reaching an elbow point at 2 iterations, providing an optimal trade-off between real-time performance and success rate. Failure reasoning proves crucial for tasks demanding high-precision manipulation, which are susceptible to environmental noise. It enhances both the success rate and the robot's ability to operate in intricate environments.

\begin{table}
\caption{Ablation experiments with VLMimic on real-world manipulation experiments. "SE" and "UE" are seen and unseen environments. Default settings are marked in \colorbox{gray!40}{gray}.}
\vspace{-5pt}
\begin{minipage}{\textwidth}
\footnotesize
\makeatletter\def\@captype{table}
\begin{subtable}[t]{0.32\textwidth}
\centering
\caption{Hierarchical representations.}
\begin{tabular}{ccc}
    \toprule
    Variants     & SE       \\
    \midrule
    Geometric constraints & 0.61       \\
    Semantic constraints   & 0.68       \\
    \rowcolor{gray!40} Hierarchical constraints     & 0.76         \\
    \bottomrule
\end{tabular}
\end{subtable}
\makeatletter\def\@captype{table}
\begin{subtable}[t]{0.32\textwidth}
\centering
\caption{Grasping learning.}
\begin{tabular}{c@{\hspace{13pt}}c@{\hspace{13pt}}c}
    \toprule
    Variants     & SE       \\
    \midrule
    Value prediction   & 0.52       \\
    Grouping (DBSCAN) & 0.59       \\
    \rowcolor{gray!40} Grouping with VLMs     & 0.76         \\
    \bottomrule
\end{tabular}
\end{subtable}
\makeatletter\def\@captype{table}
\begin{subtable}[t]{0.32\textwidth}
\centering
\caption{Number of videos.}
\begin{tabular}{c@{\hspace{4pt}}c@{\hspace{4pt}}|c@{\hspace{4pt}}c}
    \toprule
    Number     & SE & Number     & SE      \\
    \midrule
    1 &  0.68 & 7 &   0.67    \\
    3     &  0.72   & 9 &   0.78    \\
    \rowcolor{gray!40}  5     &  0.76 &  \cellcolor{white}11     &  \cellcolor{white}0.78       \\
    \bottomrule
\end{tabular}
\end{subtable}

\makeatletter\def\@captype{table}
\begin{subtable}[t]{0.32\textwidth}
\centering
\caption{Comparison strategy.}
\begin{tabular}{c@{\hspace{15pt}}c}
    \toprule
    Variants     & UE       \\
    \midrule
    Visual comparison   & 0.64         \\
    Keypoint comparison      & 0.63           \\
    \rowcolor{gray!40} Visual with keypoints     & 0.77  \\
    \bottomrule
\end{tabular}
\end{subtable}
\makeatletter\def\@captype{table}
\begin{subtable}[t]{0.32\textwidth}
\centering
\caption{Number of iterations.}
\begin{tabular}{c@{\hspace{4pt}}c@{\hspace{7pt}}|c@{\hspace{4pt}}c}
    \toprule
    Number     & UE & Number     & UE      \\
    \midrule
    0 &  0.61 & 3 &   0.75    \\
    \rowcolor{gray!40} \cellcolor{white}1     &  \cellcolor{white}0.66   & 4 &   0.77    \\
      2     &  0.73 &  5     &  0.77       \\
    \bottomrule
\end{tabular}
\end{subtable}
\makeatletter\def\@captype{table}
\begin{subtable}[t]{0.32\textwidth}
\centering
\caption{Failure reasoning.}
\begin{tabular}{c@{\hspace{4pt}}c@{\hspace{7pt}}|c@{\hspace{4pt}}c}
    \toprule
    Number     & UE & Number     & UE      \\
    \midrule
    0 &  0.71 & 3 &   0.78    \\
    1     &  0.74   & 4 &   0.77    \\
    \rowcolor{gray!40}  2     &  0.77 &  \cellcolor{white}5     &  \cellcolor{white}0.77       \\
    \bottomrule
\end{tabular}
\end{subtable}
\end{minipage}
\label{Ablation}
\vspace{-16pt}
\end{table}

\section{Conclusion} \label{Conclu}
In this paper, we present VLMimic, a novel approach that harnesses VLMs to learn skills with even fine-grained action levels from a limited number of human videos, and effectively generalize them to unseen environments. VLMimic first extracts object-centric interactions from human videos, and learns skills based on these interactions, using hierarchical constraint representations. In unseen environments, these skills are updated through an iterative comparison strategy. Extensive experiments conducted on various manipulation and challenging long-horizon tasks demonstrate the superior performance achieved by our VLMimic, utilizing only 5 human videos without requiring additional training, exhibiting strong skill acquisition and adaptation capabilities. 

\textbf{Limitations}. \textcolor{black}{Despite the promising performance exhibited by VLMimic, current VLMs are still limited by inference latency and computational resource requirements. We believe that the progression of lightweight VLMs will mitigate these limitations. We do not foresee obvious undesirable ethical or social impacts at this moment. Detailed discussions are provided in Appendix \ref{Limitations}}

{\small
\bibliographystyle{unsrt}
\bibliography{egbib}    

\begin{thebibliography}{10}

\bibitem{mandlekar2020learning}
Ajay Mandlekar, Danfei Xu, Roberto Mart{\'\i}n-Mart{\'\i}n, Silvio Savarese, and Li~Fei-Fei.
\newblock Learning to generalize across long-horizon tasks from human demonstrations.
\newblock {\em arXiv preprint arXiv:2003.06085}, 2020.

\bibitem{tung2021learning}
Albert Tung, Josiah Wong, Ajay Mandlekar, Roberto Mart{\'\i}n-Mart{\'\i}n, Yuke Zhu, Li~Fei-Fei, and Silvio Savarese.
\newblock Learning multi-arm manipulation through collaborative teleoperation.
\newblock In {\em 2021 IEEE International Conference on Robotics and Automation (ICRA)}, pages 9212--9219. IEEE, 2021.

\bibitem{zeng2021transporter}
Andy Zeng, Pete Florence, Jonathan Tompson, Stefan Welker, Jonathan Chien, Maria Attarian, Travis Armstrong, Ivan Krasin, Dan Duong, Vikas Sindhwani, et~al.
\newblock Transporter networks: Rearranging the visual world for robotic manipulation.
\newblock In {\em Conference on Robot Learning}, pages 726--747. PMLR, 2021.

\bibitem{brohan2022rt}
Anthony Brohan, Noah Brown, Justice Carbajal, Yevgen Chebotar, Joseph Dabis, Chelsea Finn, Keerthana Gopalakrishnan, Karol Hausman, Alex Herzog, Jasmine Hsu, et~al.
\newblock Rt-1: Robotics transformer for real-world control at scale.
\newblock {\em arXiv preprint arXiv:2212.06817}, 2022.

\bibitem{brohan2023rt}
Anthony Brohan, Noah Brown, Justice Carbajal, Yevgen Chebotar, Xi~Chen, Krzysztof Choromanski, Tianli Ding, Danny Driess, Avinava Dubey, Chelsea Finn, et~al.
\newblock Rt-2: Vision-language-action models transfer web knowledge to robotic control.
\newblock {\em arXiv preprint arXiv:2307.15818}, 2023.

\bibitem{padalkar2023open}
Abhishek Padalkar, Acorn Pooley, Ajinkya Jain, Alex Bewley, Alex Herzog, Alex Irpan, Alexander Khazatsky, Anant Rai, Anikait Singh, Anthony Brohan, et~al.
\newblock Open x-embodiment: Robotic learning datasets and rt-x models.
\newblock {\em arXiv preprint arXiv:2310.08864}, 2023.

\bibitem{chi2023diffusion}
Cheng Chi, Siyuan Feng, Yilun Du, Zhenjia Xu, Eric Cousineau, Benjamin Burchfiel, and Shuran Song.
\newblock Diffusion policy: Visuomotor policy learning via action diffusion.
\newblock {\em arXiv preprint arXiv:2303.04137}, 2023.

\bibitem{liang2023adaptdiffuser}
Zhixuan Liang, Yao Mu, Mingyu Ding, Fei Ni, Masayoshi Tomizuka, and Ping Luo.
\newblock Adaptdiffuser: Diffusion models as adaptive self-evolving planners.
\newblock {\em arXiv preprint arXiv:2302.01877}, 2023.

\bibitem{huo2023human}
Mingxiao Huo, Mingyu Ding, Chenfeng Xu, Thomas Tian, Xinghao Zhu, Yao Mu, Lingfeng Sun, Masayoshi Tomizuka, and Wei Zhan.
\newblock Human-oriented representation learning for robotic manipulation.
\newblock {\em arXiv preprint arXiv:2310.03023}, 2023.

\bibitem{liang2024skilldiffuser}
Zhixuan Liang, Yao Mu, Hengbo Ma, Masayoshi Tomizuka, Mingyu Ding, and Ping Luo.
\newblock Skilldiffuser: Interpretable hierarchical planning via skill abstractions in diffusion-based task execution.
\newblock In {\em Proceedings of the IEEE/CVF Conference on Computer Vision and Pattern Recognition}, pages 16467--16476, 2024.

\bibitem{grauman2022ego4d}
Kristen Grauman, Andrew Westbury, Eugene Byrne, Zachary Chavis, Antonino Furnari, Rohit Girdhar, Jackson Hamburger, Hao Jiang, Miao Liu, Xingyu Liu, et~al.
\newblock Ego4d: Around the world in 3,000 hours of egocentric video.
\newblock In {\em Proceedings of the IEEE/CVF Conference on Computer Vision and Pattern Recognition}, pages 18995--19012, 2022.

\bibitem{xiao2022masked}
Tete Xiao, Ilija Radosavovic, Trevor Darrell, and Jitendra Malik.
\newblock Masked visual pre-training for motor control.
\newblock {\em arXiv preprint arXiv:2203.06173}, 2022.

\bibitem{nair2022r3m}
Suraj Nair, Aravind Rajeswaran, Vikash Kumar, Chelsea Finn, and Abhinav Gupta.
\newblock R3m: A universal visual representation for robot manipulation.
\newblock {\em arXiv preprint arXiv:2203.12601}, 2022.

\bibitem{shaw2023videodex}
Kenneth Shaw, Shikhar Bahl, and Deepak Pathak.
\newblock Videodex: Learning dexterity from internet videos.
\newblock In {\em Conference on Robot Learning}, pages 654--665. PMLR, 2023.

\bibitem{schmeckpeper2020reinforcement}
Karl Schmeckpeper, Oleh Rybkin, Kostas Daniilidis, Sergey Levine, and Chelsea Finn.
\newblock Reinforcement learning with videos: Combining offline observations with interaction.
\newblock {\em arXiv preprint arXiv:2011.06507}, 2020.

\bibitem{edwards2019perceptual}
Ashley~D Edwards and Charles~L Isbell.
\newblock Perceptual values from observation.
\newblock {\em arXiv preprint arXiv:1905.07861}, 2019.

\bibitem{schmeckpeper2020learning}
Karl Schmeckpeper, Annie Xie, Oleh Rybkin, Stephen Tian, Kostas Daniilidis, Sergey Levine, and Chelsea Finn.
\newblock Learning predictive models from observation and interaction.
\newblock In {\em European Conference on Computer Vision}, pages 708--725. Springer, 2020.

\bibitem{chen2021learning}
Annie~S Chen, Suraj Nair, and Chelsea Finn.
\newblock Learning generalizable robotic reward functions from" in-the-wild" human videos.
\newblock {\em arXiv preprint arXiv:2103.16817}, 2021.

\bibitem{shao2021concept2robot}
Lin Shao, Toki Migimatsu, Qiang Zhang, Karen Yang, and Jeannette Bohg.
\newblock Concept2robot: Learning manipulation concepts from instructions and human demonstrations.
\newblock {\em The International Journal of Robotics Research}, 40(12-14):1419--1434, 2021.

\bibitem{zakka2022xirl}
Kevin Zakka, Andy Zeng, Pete Florence, Jonathan Tompson, Jeannette Bohg, and Debidatta Dwibedi.
\newblock Xirl: Cross-embodiment inverse reinforcement learning.
\newblock In {\em Conference on Robot Learning}, pages 537--546. PMLR, 2022.

\bibitem{bahl2022human}
Shikhar Bahl, Abhinav Gupta, and Deepak Pathak.
\newblock Human-to-robot imitation in the wild.
\newblock {\em arXiv preprint arXiv:2207.09450}, 2022.

\bibitem{sieb2020graph}
Maximilian Sieb, Zhou Xian, Audrey Huang, Oliver Kroemer, and Katerina Fragkiadaki.
\newblock Graph-structured visual imitation.
\newblock In {\em Conference on Robot Learning}, pages 979--989. PMLR, 2020.

\bibitem{sharma2019third}
Pratyusha Sharma, Deepak Pathak, and Abhinav Gupta.
\newblock Third-person visual imitation learning via decoupled hierarchical controller.
\newblock {\em Advances in Neural Information Processing Systems}, 32, 2019.

\bibitem{kumar2023graph}
Sateesh Kumar, Jonathan Zamora, Nicklas Hansen, Rishabh Jangir, and Xiaolong Wang.
\newblock Graph inverse reinforcement learning from diverse videos.
\newblock In {\em Conference on Robot Learning}, pages 55--66. PMLR, 2023.

\bibitem{smith2019avid}
Laura Smith, Nikita Dhawan, Marvin Zhang, Pieter Abbeel, and Sergey Levine.
\newblock Avid: Learning multi-stage tasks via pixel-level translation of human videos.
\newblock {\em arXiv preprint arXiv:1912.04443}, 2019.

\bibitem{liu2018imitation}
YuXuan Liu, Abhishek Gupta, Pieter Abbeel, and Sergey Levine.
\newblock Imitation from observation: Learning to imitate behaviors from raw video via context translation.
\newblock In {\em 2018 IEEE international conference on robotics and automation (ICRA)}, pages 1118--1125. IEEE, 2018.

\bibitem{xiong2021learning}
Haoyu Xiong, Quanzhou Li, Yun-Chun Chen, Homanga Bharadhwaj, Samarth Sinha, and Animesh Garg.
\newblock Learning by watching: Physical imitation of manipulation skills from human videos.
\newblock In {\em 2021 IEEE/RSJ International Conference on Intelligent Robots and Systems (IROS)}, pages 7827--7834. IEEE, 2021.

\bibitem{mees2020adversarial}
Oier Mees, Markus Merklinger, Gabriel Kalweit, and Wolfram Burgard.
\newblock Adversarial skill networks: Unsupervised robot skill learning from video.
\newblock In {\em 2020 IEEE International Conference on Robotics and Automation (ICRA)}, pages 4188--4194. IEEE, 2020.

\bibitem{jang2022bc}
Eric Jang, Alex Irpan, Mohi Khansari, Daniel Kappler, Frederik Ebert, Corey Lynch, Sergey Levine, and Chelsea Finn.
\newblock Bc-z: Zero-shot task generalization with robotic imitation learning.
\newblock In {\em Conference on Robot Learning}, pages 991--1002. PMLR, 2022.

\bibitem{mu2024embodiedgpt}
Yao Mu, Qinglong Zhang, Mengkang Hu, Wenhai Wang, Mingyu Ding, Jun Jin, Bin Wang, Jifeng Dai, Yu~Qiao, and Ping Luo.
\newblock Embodiedgpt: Vision-language pre-training via embodied chain of thought.
\newblock {\em Advances in Neural Information Processing Systems}, 36, 2024.

\bibitem{mu2024robocodex}
Yao Mu, Junting Chen, Qinglong Zhang, Shoufa Chen, Qiaojun Yu, Chongjian Ge, Runjian Chen, Zhixuan Liang, Mengkang Hu, Chaofan Tao, et~al.
\newblock Robocodex: Multimodal code generation for robotic behavior synthesis.
\newblock {\em arXiv preprint arXiv:2402.16117}, 2024.

\bibitem{chen2023human}
Guangyan Chen, Te~Cui, Tianxing Zhou, Zicai Peng, Mengxiao Hu, Meiling Wang, Yi~Yang, and Yufeng Yue.
\newblock Human demonstrations are generalizable knowledge for robots.
\newblock {\em arXiv preprint arXiv:2312.02419}, 2023.

\bibitem{wake2023gpt}
Naoki Wake, Atsushi Kanehira, Kazuhiro Sasabuchi, Jun Takamatsu, and Katsushi Ikeuchi.
\newblock Gpt-4v (ision) for robotics: Multimodal task planning from human demonstration.
\newblock {\em arXiv preprint arXiv:2311.12015}, 2023.

\bibitem{patel2023pretrained}
Dhruvesh Patel, Hamid Eghbalzadeh, Nitin Kamra, Michael~Louis Iuzzolino, Unnat Jain, and Ruta Desai.
\newblock Pretrained language models as visual planners for human assistance.
\newblock In {\em Proceedings of the IEEE/CVF International Conference on Computer Vision}, pages 15302--15314, 2023.

\bibitem{weng2024longvlm}
Yuetian Weng, Mingfei Han, Haoyu He, Xiaojun Chang, and Bohan Zhuang.
\newblock Longvlm: Efficient long video understanding via large language models.
\newblock {\em arXiv preprint arXiv:2404.03384}, 2024.

\bibitem{li2023videochat}
KunChang Li, Yinan He, Yi~Wang, Yizhuo Li, Wenhai Wang, Ping Luo, Yali Wang, Limin Wang, and Yu~Qiao.
\newblock Videochat: Chat-centric video understanding.
\newblock {\em arXiv preprint arXiv:2305.06355}, 2023.

\bibitem{wang2024demo2code}
Yuki Wang, Gonzalo Gonzalez-Pumariega, Yash Sharma, and Sanjiban Choudhury.
\newblock Demo2code: From summarizing demonstrations to synthesizing code via extended chain-of-thought.
\newblock {\em Advances in Neural Information Processing Systems}, 36, 2024.

\bibitem{sha2023languagempc}
Hao Sha, Yao Mu, Yuxuan Jiang, Li~Chen, Chenfeng Xu, Ping Luo, Shengbo~Eben Li, Masayoshi Tomizuka, Wei Zhan, and Mingyu Ding.
\newblock Languagempc: Large language models as decision makers for autonomous driving.
\newblock {\em arXiv preprint arXiv:2310.03026}, 2023.

\bibitem{hu2023tree}
Mengkang Hu, Yao Mu, Xinmiao Yu, Mingyu Ding, Shiguang Wu, Wenqi Shao, Qiguang Chen, Bin Wang, Yu~Qiao, and Ping Luo.
\newblock Tree-planner: Efficient close-loop task planning with large language models.
\newblock {\em arXiv preprint arXiv:2310.08582}, 2023.

\bibitem{gao2024dag}
Zeyu Gao, Yao Mu, Jinye Qu, Mengkang Hu, Lingyue Guo, Ping Luo, and Yanfeng Lu.
\newblock Dag-plan: Generating directed acyclic dependency graphs for dual-arm cooperative planning.
\newblock {\em arXiv preprint arXiv:2406.09953}, 2024.

\bibitem{das2021model}
Neha Das, Sarah Bechtle, Todor Davchev, Dinesh Jayaraman, Akshara Rai, and Franziska Meier.
\newblock Model-based inverse reinforcement learning from visual demonstrations.
\newblock In {\em Conference on Robot Learning}, pages 1930--1942. PMLR, 2021.

\bibitem{zhang2023recognize}
Youcai Zhang, Xinyu Huang, Jinyu Ma, Zhaoyang Li, Zhaochuan Luo, Yanchun Xie, Yuzhuo Qin, Tong Luo, Yaqian Li, Shilong Liu, et~al.
\newblock Recognize anything: A strong image tagging model.
\newblock {\em arXiv preprint arXiv:2306.03514}, 2023.

\bibitem{wang2023caption}
Teng Wang, Jinrui Zhang, Junjie Fei, Yixiao Ge, Hao Zheng, Yunlong Tang, Zhe Li, Mingqi Gao, Shanshan Zhao, Ying Shan, et~al.
\newblock Caption anything: Interactive image description with diverse multimodal controls.
\newblock {\em arXiv preprint arXiv:2305.02677}, 2023.

\bibitem{pan2023tokenize}
Ting Pan, Lulu Tang, Xinlong Wang, and Shiguang Shan.
\newblock Tokenize anything via prompting.
\newblock {\em arXiv preprint arXiv:2312.09128}, 2023.

\bibitem{liu2023grounding}
Shilong Liu, Zhaoyang Zeng, Tianhe Ren, Feng Li, Hao Zhang, Jie Yang, Chunyuan Li, Jianwei Yang, Hang Su, Jun Zhu, et~al.
\newblock Grounding dino: Marrying dino with grounded pre-training for open-set object detection.
\newblock {\em arXiv preprint arXiv:2303.05499}, 2023.

\bibitem{yang2021aot}
Zongxin Yang, Yunchao Wei, and Yi~Yang.
\newblock Associating objects with transformers for video object segmentation.
\newblock In {\em Advances in Neural Information Processing Systems (NeurIPS)}, 2021.

\bibitem{yang2022deaot}
Zongxin Yang and Yi~Yang.
\newblock Decoupling features in hierarchical propagation for video object segmentation.
\newblock In {\em Advances in Neural Information Processing Systems (NeurIPS)}, 2022.

\bibitem{kirillov2023segment}
Alexander Kirillov, Eric Mintun, Nikhila Ravi, Hanzi Mao, Chloe Rolland, Laura Gustafson, Tete Xiao, Spencer Whitehead, Alexander~C Berg, Wan-Yen Lo, et~al.
\newblock Segment anything.
\newblock {\em arXiv preprint arXiv:2304.02643}, 2023.

\bibitem{cheng2023segment}
Yangming Cheng, Liulei Li, Yuanyou Xu, Xiaodi Li, Zongxin Yang, Wenguan Wang, and Yi~Yang.
\newblock Segment and track anything.
\newblock {\em arXiv preprint arXiv:2305.06558}, 2023.

\bibitem{rong2020frankmocap}
Yu~Rong, Takaaki Shiratori, and Hanbyul Joo.
\newblock Frankmocap: Fast monocular 3d hand and body motion capture by regression and integration.
\newblock {\em arXiv preprint arXiv:2008.08324}, 2020.

\bibitem{besl1992method}
Paul~J Besl and Neil~D McKay.
\newblock Method for registration of 3-d shapes.
\newblock In {\em Sensor fusion IV: control paradigms and data structures}, volume 1611, pages 586--606. Spie, 1992.

\bibitem{rusinkiewicz2001efficient}
Szymon Rusinkiewicz and Marc Levoy.
\newblock Efficient variants of the icp algorithm.
\newblock In {\em Proceedings third international conference on 3-D digital imaging and modeling}, pages 145--152. IEEE, 2001.

\bibitem{wen2023bundlesdf}
Bowen Wen, Jonathan Tremblay, Valts Blukis, Stephen Tyree, Thomas M{\"u}ller, Alex Evans, Dieter Fox, Jan Kautz, and Stan Birchfield.
\newblock Bundlesdf: Neural 6-dof tracking and 3d reconstruction of unknown objects.
\newblock In {\em Proceedings of the IEEE/CVF Conference on Computer Vision and Pattern Recognition}, pages 606--617, 2023.

\bibitem{wen2023foundationpose}
Bowen Wen, Wei Yang, Jan Kautz, and Stan Birchfield.
\newblock Foundationpose: Unified 6d pose estimation and tracking of novel objects.
\newblock {\em arXiv preprint arXiv:2312.08344}, 2023.

\bibitem{wang2023prompt}
Yen-Jen Wang, Bike Zhang, Jianyu Chen, and Koushil Sreenath.
\newblock Prompt a robot to walk with large language models.
\newblock {\em arXiv preprint arXiv:2309.09969}, 2023.

\bibitem{mirchandani2023large}
Suvir Mirchandani, Fei Xia, Pete Florence, Brian Ichter, Danny Driess, Montserrat~Gonzalez Arenas, Kanishka Rao, Dorsa Sadigh, and Andy Zeng.
\newblock Large language models as general pattern machines.
\newblock {\em arXiv preprint arXiv:2307.04721}, 2023.

\bibitem{berenson2011task}
Dmitry Berenson, Siddhartha Srinivasa, and James Kuffner.
\newblock Task space regions: A framework for pose-constrained manipulation planning.
\newblock {\em The International Journal of Robotics Research}, 30(12):1435--1460, 2011.

\bibitem{sucan2012open}
Ioan~A Sucan, Mark Moll, and Lydia~E Kavraki.
\newblock The open motion planning library.
\newblock {\em IEEE Robotics \& Automation Magazine}, 19(4):72--82, 2012.

\bibitem{skreta2024replan}
Marta Skreta, Zihan Zhou, Jia~Lin Yuan, Kourosh Darvish, Al{\'a}n Aspuru-Guzik, and Animesh Garg.
\newblock Replan: Robotic replanning with perception and language models.
\newblock {\em arXiv preprint arXiv:2401.04157}, 2024.

\bibitem{hu2023look}
Yingdong Hu, Fanqi Lin, Tong Zhang, Li~Yi, and Yang Gao.
\newblock Look before you leap: Unveiling the power of gpt-4v in robotic vision-language planning.
\newblock {\em arXiv preprint arXiv:2311.17842}, 2023.

\bibitem{fang2020graspnet}
Hao-Shu Fang, Chenxi Wang, Minghao Gou, and Cewu Lu.
\newblock Graspnet-1billion: A large-scale benchmark for general object grasping.
\newblock In {\em Proceedings of the IEEE/CVF conference on computer vision and pattern recognition}, pages 11444--11453, 2020.

\bibitem{fang2023anygrasp}
Hao-Shu Fang, Chenxi Wang, Hongjie Fang, Minghao Gou, Jirong Liu, Hengxu Yan, Wenhai Liu, Yichen Xie, and Cewu Lu.
\newblock Anygrasp: Robust and efficient grasp perception in spatial and temporal domains.
\newblock {\em IEEE Transactions on Robotics}, 2023.

\bibitem{ju2024robo}
Yuanchen Ju, Kaizhe Hu, Guowei Zhang, Gu~Zhang, Mingrun Jiang, and Huazhe Xu.
\newblock Robo-abc: Affordance generalization beyond categories via semantic correspondence for robot manipulation.
\newblock {\em arXiv preprint arXiv:2401.07487}, 2024.

\bibitem{liang2023code}
Jacky Liang, Wenlong Huang, Fei Xia, Peng Xu, Karol Hausman, Brian Ichter, Pete Florence, and Andy Zeng.
\newblock Code as policies: Language model programs for embodied control.
\newblock In {\em 2023 IEEE International Conference on Robotics and Automation (ICRA)}, pages 9493--9500. IEEE, 2023.

\bibitem{james2020rlbench}
Stephen James, Zicong Ma, David~Rovick Arrojo, and Andrew~J Davison.
\newblock Rlbench: The robot learning benchmark \& learning environment.
\newblock {\em IEEE Robotics and Automation Letters}, 5(2):3019--3026, 2020.

\bibitem{ahn2022can}
Michael Ahn, Anthony Brohan, Noah Brown, Yevgen Chebotar, Omar Cortes, Byron David, Chelsea Finn, Chuyuan Fu, Keerthana Gopalakrishnan, Karol Hausman, et~al.
\newblock Do as i can, not as i say: Grounding language in robotic affordances.
\newblock {\em arXiv preprint arXiv:2204.01691}, 2022.

\bibitem{xiao2022robotic}
Ted Xiao, Harris Chan, Pierre Sermanet, Ayzaan Wahid, Anthony Brohan, Karol Hausman, Sergey Levine, and Jonathan Tompson.
\newblock Robotic skill acquisition via instruction augmentation with vision-language models.
\newblock {\em arXiv preprint arXiv:2211.11736}, 2022.

\bibitem{yu2023scaling}
Tianhe Yu, Ted Xiao, Austin Stone, Jonathan Tompson, Anthony Brohan, Su~Wang, Jaspiar Singh, Clayton Tan, Jodilyn Peralta, Brian Ichter, et~al.
\newblock Scaling robot learning with semantically imagined experience.
\newblock {\em arXiv preprint arXiv:2302.11550}, 2023.

\end{thebibliography}
}

\end{document}